\documentclass[runningheads]{llncs}

\usepackage{accv}            

\usepackage{accvabbrv}

\usepackage{graphicx}
\usepackage{booktabs}
\usepackage{multirow}   
\usepackage{pifont}     
\usepackage{textcomp}   
\usepackage{array}      

\usepackage{hyperref}        
\usepackage{orcidlink}

\newcommand{\new}[1]{#1}

\begin{document}

\title{Fine-grained Motion Retrieval via Joint-Angle Motion Images and Token-Patch Late Interaction}
\titlerunning{Fine-grained Motion Retrieval via Joint-Angle Motion Images}
\author{Yao Zhang\inst{1} \and
Zhuchenyang Liu\inst{1} \and
Yalan He\inst{2} \and
Thomas Ploetz\inst{3} \and
Yu Xiao\inst{1}}
\authorrunning{Y. Zhang et al.}
\institute{Aalto University, Espoo, Finland\\
\email{\{yao.1.zhang,zhuchenyang.liu,yu.xiao\}@aalto.fi} \and
Fudan University, Shanghai, China\\
\email{23307130480@m.fudan.edu.cn} \and
Georgia Institute of Technology, Atlanta GA, USA\\
\email{thomas.ploetz@gatech.edu}}

\maketitle

\begin{abstract}
Text-motion retrieval aims to learn a semantically aligned latent space between natural language descriptions and 3D human motion skeleton sequences, enabling bidirectional search across the two modalities. Most existing methods use a dual-encoder framework that compresses motion and text into global embeddings, discarding fine-grained local correspondences, and thus reducing accuracy. Additionally, these global-embedding methods offer limited interpretability of the retrieval results. To overcome these limitations, we propose an interpretable, joint-angle-based motion representation that maps joint-level local features into a structured pseudo-image, compatible with pre-trained Vision Transformers. For text-to-motion retrieval, we employ MaxSim, a token-wise late interaction mechanism, and enhance it with Masked Language Modeling regularization to foster robust, interpretable text-motion alignment. Extensive experiments on HumanML3D and KIT-ML show that our method outperforms state-of-the-art text-motion retrieval approaches while offering interpretable fine-grained correspondences between text and motion. Code and models are available at: https://github.com/yaozhang182/MotionColBert.
\end{abstract}

\section{Introduction}
\label{sec:intro}

Text-motion retrieval is a pivotal task in human motion understanding, whose core objective is to establish a semantic-aligned latent space between 3D human motion skeleton sequences and natural language descriptions~\cite{guo2022generating,petrovich2023tmr,plappert2016kit}. A well-aligned latent space facilitates bidirectional retrieval, where the accuracy and granularity of alignment significantly impacts the performance of downstream tasks such as text-driven motion generation~\cite{tevet2022human}, motion captioning~\cite{guo2022tm2t,jiang2024motiongpt}, and language-guided motion editing~\cite{zhang2024motiondiffuse}.

Recent efforts aim to achieve accurate and fine-grained alignment from various perspectives. On the motion representation front, some studies~\cite{yu2024exploring, sgar2025} have converted motion skeleton sequences into pseudo-images to leverage pre-trained vision models. However, these representations, derived directly from raw 3D joint positions, conflate global translational movement with individual joint movements, hindering the distinction of subtle kinematic differences (see Sec.~\ref{sec:motion_rep}). For text modality, previous works have employed large language models for generating augmented or part-level descriptions~\cite{kinmo2025,sgar2025}, or developed complex cross-modal attention modules for finer alignment~\cite{cmmm2025,secl2025}. Nonetheless, these solutions introduce significant external dependencies and computational overhead. Furthermore, most existing methods, such as TMR~\cite{petrovich2023tmr}, adhere to a global-embedding alignment paradigm by encoding the input sequences into single global vectors for alignment. Although computationally efficient, this global-embedding paradigm compresses the rich content of a motion sequence and its textual description into a single vector, inevitably discarding fine-grained local information. As a result, it restricts retrieval performance and complicates the alignment of specific text tokens with precise body movements, essential for distinguishing similar motions.

This paper tackles these challenges from two key perspectives. First, we address the issue of conflating global and local joint movements by introducing a joint-angle-based motion representation. Anatomically defined joint angles~\cite{wu2002isb, wu2005isb} are translation-invariant and encode each joint independently, explicitly decoupling the body's global trajectory from local joint movement (detailed in Sec.~\ref{sec:related_repr}). By projecting these decoupled features, rather than raw 3D joint positions, into a structured $224 \times 224$ pseudo-image, we effectively leverage visual priors from pre-trained Vision Transformers (ViT). This approach ensures that distinct spatial regions correspond to specific joints, providing a natural basis for part-level alignment.

Second, inspired by late interaction mechanisms effective in text retrieval~\cite{khattab2020colbert} and visual document retrieval~\cite{faysse2024colpali}, we replace the global-embedding alignment with a Maximum Similarity (MaxSim) operator for explicit token-to-patch matching. This operator retains the maximum similarity score for each text token across all motion patches, producing interpretable, fine-grained correspondence maps. However, since MaxSim computes similarity at the individual token level, a key challenge arises: each token and patch embedding must carry sufficient contextual information to support reliable matching. On the motion side, sequences are often dominated by static or transitional poses, which may lead MaxSim to assign high similarity to uninformative patches that happen to be feature-similar to a query token. On the text side, semantically vacuous tokens such as ``a'' or ``person'' can act as noise anchors, matching to arbitrary motion patches and diluting the alignment signal; even content words like ``hand'' may match unrelated patches if they lack sentence-level context. To address these, we introduce Masked Language Modeling (MLM)~\cite{devlin2019bert} as regularization, training the text encoder to reconstruct masked tokens from their surrounding context. This innovative approach produces contextually enriched embeddings that stabilize fine-grained matching, all without the need for data augmentation or external models. To the best of our knowledge as discussed in Sec.~\ref{sec:tmr} and~\ref{sec:late_interaction}, this is the first work to replace the global-embedding paradigm in text-motion retrieval with a structurally grounded, fine-grained late interaction mechanism.

Extensive experiments on HumanML3D and KIT-ML demonstrate that our approach, which combines joint-angle-based motion representation with token-patch late interaction, consistently surpasses state-of-the-art text--motion retrieval baselines. Additionally, it produces interpretable correspondence maps that transparently align textual semantics with specific body joints and temporal phases. Details of our methods and evaluation results are presented in Secs.~\ref{sec:methodology} and~\ref{sec:experiments}, respectively, followed by the conclusion in Sec.~\ref{sec:conclusion}.

\section{Background and Related Work}
\label{sec:related}
This section reviews existing human motion representations (Sec.~\ref{sec:related_repr}), text-motion retrieval approaches (Sec.~\ref{sec:tmr}), and late interaction mechanisms used in text or document retrieval (Sec.~\ref{sec:late_interaction}).

\subsection{Human Motion Representations}
\label{sec:related_repr}

3D human motion sequences are typically represented as frame-wise skeletal features. The most widely adopted format is the 263-dimensional feature vector proposed by Guo et al.~\cite{guo2022generating}, which concatenates, per frame, the root angular velocity, root linear velocity, joint positions, 6D joint rotations, joint velocities, and binary foot-contact labels. Recent motion-generation studies revisit the choice of motion representation: MARDM~\cite{mardm2025} analyzes the redundancy of such multi-view pose encodings, while ACMDM~\cite{acmdm2025} shows that directly using absolute joint coordinates can ease motion generation. For retrieval tasks, a compact and discriminative representation is also preferrable. MoPatch~\cite{yu2024exploring} took a step forward by converting motion sequences into image-like patches and leveraging pre-trained ViT encoders, demonstrating that visual priors from ImageNet can alleviate data scarcity in motion understanding. However, its motion images are constructed directly from raw 3D joint positions grouped by body region, obscuring fine-grained kinematic differences between joints.

In biomechanics, the Joint Coordinate Systems (JCS) standardized by the International Society of Biomechanics (ISB) decompose each joint's motion into clinically meaningful components, such as flexion/extension, abduction/adduction, and internal/external rotation, along well-defined anatomical planes~\cite{wu2002isb,wu2005isb}. These angles offer anatomical interpretability, with each degree of freedom (DoF) corresponding to a specific type of joint articulation~\cite{schlegel2024joint}. They are also inherently translation-invariant, describing how a joint moves relative to its parent segment (\eg, the thigh is the parent segment of the knee joint, and the pelvis is the parent of the hip). Despite these benefits, joint angles remain underexplored for cross-modal motion retrieval. We address this by constructing a structured Motion Image from per-joint angular features, with each spatial region encoding a distinct joint to enable fine-grained token-to-patch alignment.

\subsection{Text-Motion Retrieval}
\label{sec:tmr}

Cross-modal retrieval between natural language and 3D human skeleton motion has attracted growing attention~\cite{guo2022generating,plappert2016kit,mahmood2019amass}. TMR~\cite{petrovich2023tmr}, inspired by the architecture of image-text~\cite{radford2021learning} and video-text retrieval models~\cite{xu2021videoclip}, introduced a dual-encoder framework for text-motion retrieval in which a motion encoder and a text encoder independently map their inputs into a shared latent space, then the Text-to-Motion (T2M) and Motion-to-Text (M2T) retrieval are performed via cosine similarity between the resulting global embeddings. Using a similar architecture as TMR~\cite{petrovich2023tmr}, several works seek finer-grained alignment by either leveraging LLM to enrich the training data (\eg, CAR~\cite{car2024}, SGAR~\cite{sgar2025}, and KinMo~\cite{kinmo2025}) or introducing coarser notions of locality (\eg, Lyu et al.~\cite{cmmm2025}, ReMoGPT~\cite{remogpt2025}, SECL~\cite{secl2025}); more recent efforts pursue part-specific or hierarchical fine-grained matching (\eg, WaMo~\cite{wamo2025}, Chen et al.~\cite{beyondglobal2026}). Recent works also target multi-modal retrieval that aligns motion with additional modalities such as audio or images to learn a broader joint embedding space (\eg, MotionBind~\cite{motionbind2025}, MonSTeR~\cite{monster2025}, and Multi-Modal Motion Retrieval~\cite{trimodal2024, multimodalmr2026}). However, none of these establishes direct, structurally grounded correspondence between individual words and specific body regions or temporal segments, precisely the capability our token-to-patch alignment provides.

\subsection{Late Interaction in Retrieval}
\label{sec:late_interaction}

Late interaction was introduced by ColBERT~\cite{khattab2020colbert} for text retrieval. Unlike single-vector models that compress documents into one embedding, ColBERT retains per-token representations and computes relevance via a Maximum Similarity (MaxSim) operator: each text token in the query finds its best-matching document token, and the final score aggregates these maxima. This preserves fine-grained detail while retaining the computational efficiency of independent encoding. Specifically, unlike cross-attention mechanisms~\cite{devlin2019bert,secl2025} that require joint processing of the query and every gallery item through the model at runtime, late interaction allows all gallery items to be pre-encoded offline. During inference, only the query needs a forward pass, and retrieval is achieved via lightweight similarity computations. The paradigm has since been extended to multi-modal settings: ColPali~\cite{faysse2024colpali} applies it to visual document retrieval, matching query text tokens against image patch embeddings without requiring OCR. These developments show that late interaction naturally suits cross-modal tasks where fine-grained correspondences exist between modality elements, however, it has not been explored in the motion domain. To the best of our knowledge, our work is the first to introduce this mechanism, enhanced with MLM-based regularization, for text-motion retrieval.

\section{Methodology}
\label{sec:methodology}

\begin{figure}[t]
    \centering
    \includegraphics[width=\linewidth]{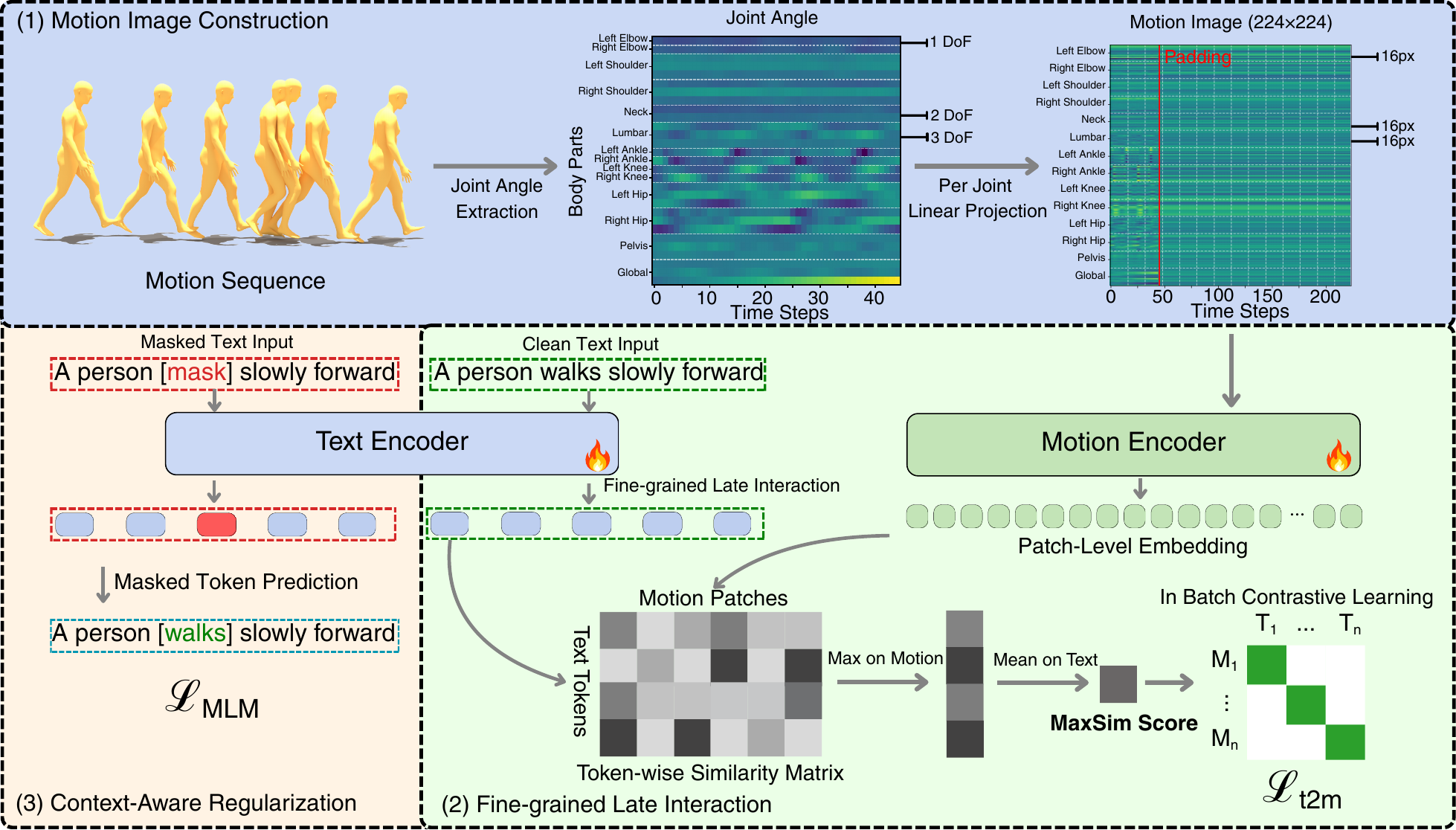}
    \caption{Overview of the three-stage training pipeline.}
    \label{fig:architecture}
\end{figure}

To enable interpretable fine-grained text-motion retrieval, we propose a framework designed to overcome the information bottleneck of traditional global embeddings. As illustrated in Fig.~\ref{fig:architecture}, given a 3D skeletal motion sequence (represented as per-frame joint positions from SMPL~\cite{guo2022generating} or similar skeletons) and a natural language description, our pipeline operates in three steps.

First, to explicitly decouple local joint movements from global body trajectories, a raw motion sequence is converted into a structured $224 \times 224$ pseudo-image, termed theMotion Image. This is achieved by extracting joint angle features via inverse kinematics and projecting each joint's Degrees of Freedom (DoF) into a uniform 16-pixel horizontal band, so that each spatial region encodes a distinct joint (Sec.~\ref{sec:motion_rep}).

Second, the Motion Image is fed into a Vision Transformer, while the text description is processed by a Transformer-based language model. They output patch-level and token-level embeddings, respectively, preserving spatiotemporal nuances for downstream matching instead of collapsing everything into a single global vector (Sec.~\ref{sec:architecture}). These embeddings are combined through a MaxSim late interaction mechanism~\cite{khattab2020colbert} to yield a fine-grained MaxSim Score for text-motion alignment (Sec.~\ref{sec:maxsim}). This score serves as the unified retrieval metric for both training and inference: during training, it is used within in-batch contrastive learning ($\mathcal{L}_{t2m}$); during inference, it directly ranks candidate motions (or texts) for retrieval.

Third, because token-level matching is sensitive to semantic noise, it is paired with a MLM auxiliary objective that enriches token embeddings with sentence-level context to stabilize the alignment (Sec.~\ref{sec:mlm}). MLM has been widely used alongside contrastive objectives in vision-language pre-training~\cite{li2021align,bao2022vlmo} to improve text representation quality; in our framework, it serves the targeted purpose of ensuring that each token embedding encodes not just lexical identity but also its role within the broader sentence, providing a stable foundation for fine-grained cross-modal matching.

The third step is used only during training; the others are performed in both training and inference. Regarding inference, for T2M retrieval, given a text query $\mathcal{T}$ and a gallery of $G$ candidate motions $\{\mathcal{M}_g\}_{g=1}^{G}$, the text encoder and motion encoder independently produce token-level and patch-level embeddings, respectively. The MaxSim Score $\mathrm{Sim}(\mathcal{T}, \mathcal{M}_g)$ is computed for each candidate, and motions are ranked accordingly. For M2T retrieval, the roles are reversed: a motion query is scored against all candidate texts using the transposed similarity matrix. Since the two encoders are independent, candidate embeddings need only be computed once and can be reused across multiple queries.

\subsection{Joint-angle-based Motion Representation}
\label{sec:motion_rep}
Joint angles describe how a joint bends or rotates relative to its parent limb segment, regardless of where the body is located in world coordinates. Unlike previous methods~\cite{cmmm2025, petrovich2023tmr, yu2024exploring} that rely on raw joint positions, we adopt a joint-angle-based motion representation to explicitly decouple local joint movement from global movement to construct fine-grained Motion Image. This provides three benefits for retrieval: (1)~each spatial band in the Motion Image encodes a distinct joint's behavior independently, enabling part-level alignment; (2)~the representation is naturally robust to global translation and heading variation; and (3)~discriminative kinematic patterns (\eg, the periodic hip flexion during walking) are preserved without being masked by smooth global drift. This angle-based representation is also invertible; we demonstrate the high-fidelity recovery of 3D joint positions via forward kinematics in Supplementary Material~1.4.

As illustrated in Fig.~\ref{fig:joint_image}(a), we decompose a motion into $K{=}14$ distinct kinematic joints (\eg, hips, knees, shoulders). Following the anatomical DoF definitions for each joint (\eg, 3 dimensions for ball-and-socket joints such as hips, 1 dimension for hinge joints such as knees) and standard medical conventions for joint angle definitions~\cite{wu2002isb,wu2005isb}, we compute per-part angular features as well as global position from the raw joint positions via an inverse kinematics pipeline. This yields an input vector $p_t \in \mathbb{R}^{29}$ at each time step $t$. Table~\ref{tab:joint_angles_summary_main} summarizes the computed joint angle features across all 14 joints.

\begin{table}[t]
\centering
\caption{Summary of joint angle features. We compute 29 kinematic dimensions corresponding to 14 joints.}
\label{tab:joint_angles_summary_main}
\resizebox{0.9\columnwidth}{!}{%
\begin{tabular}{llcl}
\toprule
\textbf{Joint} & \textbf{Features} & \textbf{DoF} & \textbf{Joint Type} \\
\midrule
Pelvis (orientation) & tilt, list, rotation & 3 & Facing direction \\
Pelvis (translation) & tx, ty, tz & 3 & Global position \\
\midrule
Hips ($\times$2) & flexion, adduction, rotation & 6 & Ball-and-socket \\
Knees ($\times$2) & bending & 2 & Hinge \\
Ankles ($\times$2) & bending & 2 & Hinge \\
Lumbar & extension, bending, rotation & 3 & Spinal (3-DoF) \\
Shoulders ($\times$2) & flexion, adduction, rotation & 6 & Ball-and-socket \\
Elbows ($\times$2) & flexion & 2 & Hinge \\
Neck & flexion, adduction & 2 & Spinal (2-DoF) \\
\midrule
Total  & & 29 & \\
\bottomrule
\end{tabular}%
}
\end{table}

\begin{figure}[t]
    \centering
    \includegraphics[width=\columnwidth]{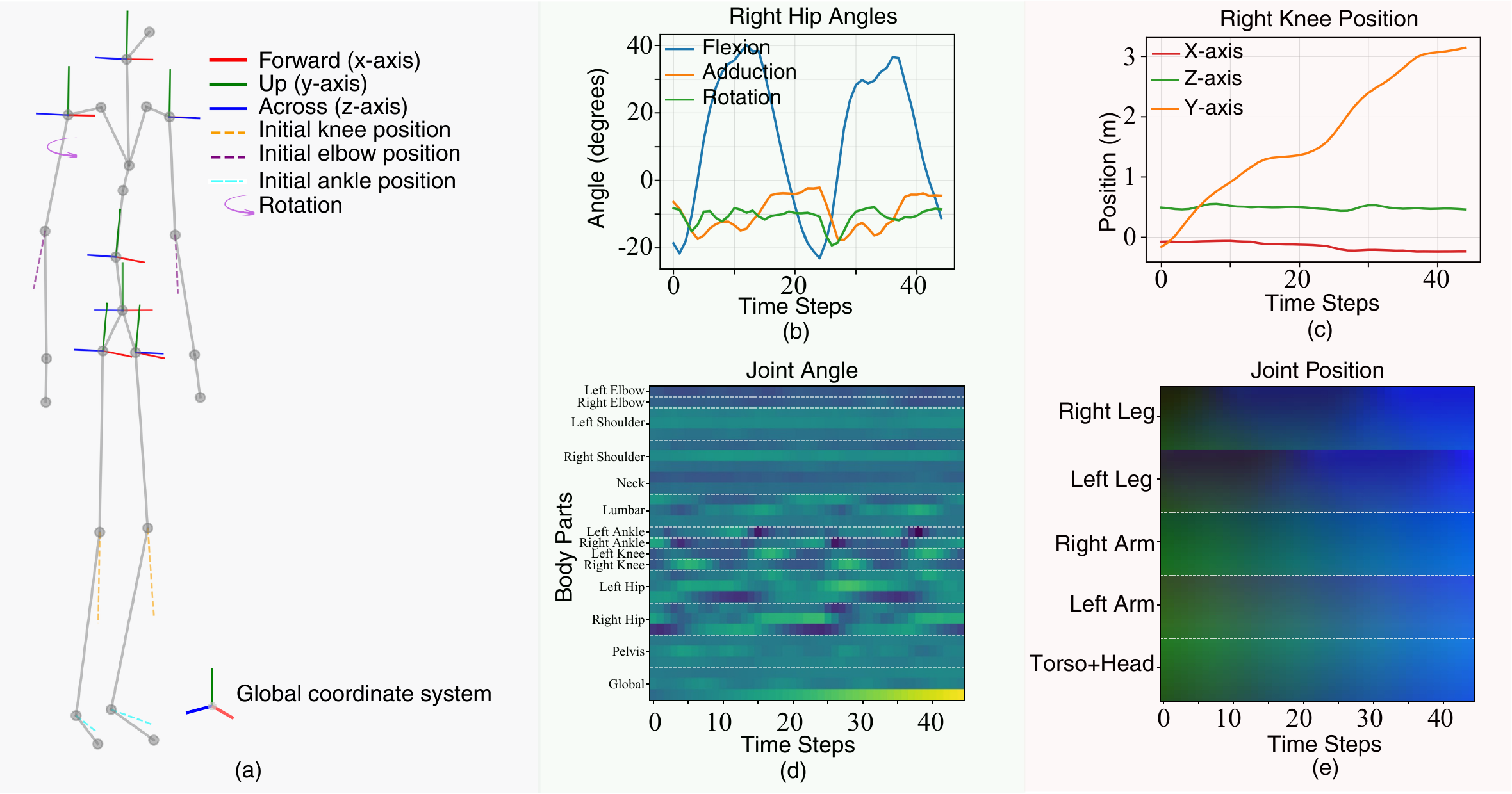}
    \caption{Joint-angle vs.\ joint-position-based representations for ``a person walks slowly forward''. (a)~Skeletal structure and body-centric axes. (b)~Right hip angles. (c)~Right knee positions. (d)~Our 29-dimension joint angle Motion Image: each band encodes a distinct joint. (e)~MoPatch~\cite{yu2024exploring} position image.}
    \label{fig:joint_image}
\end{figure}

\paragraph{Joint angle extraction.}
We construct a body-centric coordinate system $R_{\rm body}$ anchored to the pelvis at each frame (Supplementary Material~1.1), and extract joint angles via inverse kinematics depending on joint type. For ball-and-socket joints (hips, shoulders; 3-DoF), we transform the child position into the parent's local frame to obtain $\tilde{v} \in \mathbb{R}^3$ and decompose it into flexion/extension and adduction/abduction:
\begin{equation}
    \theta_{\rm flex} = {\rm atan2}(\tilde{v}_x,\, -\tilde{v}_y), \quad
    \theta_{\rm add} = {\rm sign}(\tilde{v}_z)\cdot\arccos\!\left(
    \frac{\tilde{v} \cdot \tilde{v}_{xy}}
    {\|\tilde{v}\|\,\|\tilde{v}_{xy}\|}\right)
\end{equation}
where $\tilde{v}_{xy}$ is the sagittal-plane projection of $\tilde{v}$. The axial rotation $\theta_{\rm rot}$ is obtained by propagating the parent frame and measuring the grandchild twist (Supplementary Material~1.2). Hinge joints (1-DoF) use the angle between adjacent limb segments, and spinal joints (2--3 DoF) follow the same decomposition but reference the upward $y$-axis. All frames follow hierarchical recursive propagation, ensuring translation-invariant features (full derivation in Supplementary Material~1).

\paragraph{Motion Image construction.}
Each motion is split into joint-specific features $\{p_{t,k}\}_{k=1}^{K}$. Since joints have varying DoF (1--3 dimensions), we use learnable linear projections $\{\phi_k\}_{k=1}^{K}$ to map them into a unified space:
\begin{equation}
    h_{t,k} = \phi_k(p_{t,k}) \in \mathbb{R}^{d_{part}}
\end{equation}
where $d_{part}{=}16$ matches the ViT~\cite{dosovitskiy2020image} patch size so that each joint occupies exactly one 16-pixel horizontal band, yielding a one-to-one joint-to-patch mapping (adjustable to other ViT configurations, \eg, $d_{part}{=}14$). At each frame $t$, the projected features of all $K$ joints are concatenated as one column $H_t \in \mathbb{R}^{D_{emb}}$ ($D_{emb} = K \times d_{part} = 224$). Stacking all frames along the temporal axis and padding to $T{=}224$ produces a pseudo-image $I_{motion} \in \mathbb{R}^{224 \times 224}$ termed the ``Motion Image''. As shown in Fig.~\ref{fig:architecture}, each horizontal band corresponds to a specific joint, naturally supporting part-level alignment with pre-trained ViTs.

Fig.~\ref{fig:joint_image} illustrates the translation-invariant features using the motion ``a person walks slowly forward'' as an example. The right hip's joint angles (Fig.~\ref{fig:joint_image}(b)) exhibit clear periodic gait patterns, while the corresponding joint positions (Fig.~\ref{fig:joint_image}(c)) are dominated by global trajectory drift. At the whole-body level, the joint angle Motion Image (Fig.~\ref{fig:joint_image}(d)) shows temporally localized, joint-specific activations, whereas the position-based image (Fig.~\ref{fig:joint_image}(e)) shows uniform drift across all bands, obscuring kinematic differences.

\subsection{Dual-Stream Architecture}
\label{sec:architecture}
The framework consists of a dual-stream architecture comprising a Motion Encoder and a Text Encoder, designed to extract dense feature representations for the subsequent late interaction.

\paragraph{Motion Encoder.}
The motion encoder $\mathcal{E}_m$ uses a ViT backbone. The single-channel Motion Image $I_{motion}$ is repeated across RGB channels to match the ImageNet-pre-trained input format, yielding patch embeddings:
\begin{equation}
    V = \mathcal{E}_m(I_{motion}) = \{v_1, v_2, ..., v_{N}\} \in \mathbb{R}^{N \times d}
\end{equation}
where $N$ is the number of patches (\eg, 196 for a $14 \times 14$ grid) and $d$ is the feature dimension. Unlike CLIP-style models, we retain the full patch sequence $V$ rather than pooling into a global \texttt{[CLS]} token.

\paragraph{Text Encoder.}
The text encoder $\mathcal{E}_t$ (a Transformer-based language model, \eg, DistilBERT) maps a description $\mathcal{T}$ to token-level hidden states:
\begin{equation}
    L = \mathcal{E}_t(\mathcal{T}) = \{l_1, l_2, ..., l_{M}\} \in \mathbb{R}^{M \times d}
\end{equation}
where $M$ is the sequence length. We use the content-token embeddings $\{l_i\}$ (excluding \texttt{[CLS]} and \texttt{[SEP]}) for fine-grained interaction.

\subsection{Fine-Grained Late Interaction (MaxSim)}
\label{sec:maxsim}

While standard global pooling aggregates information prematurely, MaxSim explicitly models the alignment between distinct text tokens and motion patches. We define the token-patch interaction matrix $S \in \mathbb{R}^{M \times N}$ as the dot product between normalized text and motion features:
\begin{equation}
\label{eq:maxsim_matrix}
    S_{ij} = \frac{l_i \cdot v_j^T}{\|l_i\| \|v_j\|}
\end{equation}
We compute MaxSim in the T2M direction: for each text token $l_i$ (\eg, representing ``hand''), we identify the motion patch $v_j$ that yields the maximum activation response. The similarity score for a text-motion pair is then obtained by averaging these per-token maxima:
\begin{equation}
\label{eq:maxsim}
    \text{Sim}(\mathcal{T}, \mathcal{M}) = \frac{1}{M} \sum_{i=1}^{M} \max_{j=1}^{N} (S_{ij})
\end{equation}
This mechanism allows the model to dynamically ground each word to its most relevant kinematic feature, filtering out irrelevant motion background. We compute MaxSim only in the T2M direction because motion sequences contain far more patches ($N{=}196$) than text tokens ($M \ll N$), and many patches encode static or repetitive postures with no textual counterpart; a reverse max-over-text would introduce substantial noise.

\subsection{Context-Aware Regularization via MLM}
\label{sec:mlm}

To ensure that token embeddings carry sufficient contextual information for reliable MaxSim matching, we introduce MLM~\cite{devlin2019bert, li2021align} as a context-aware regularization task. During training, we randomly mask a proportion of the input tokens (\eg, 15\%) to generate a corrupted text $\mathcal{T}_{masked}$. The Text Encoder is then required to reconstruct the original tokens based on the contextual dependency of the visible tokens. This auxiliary task forces the encoder to deeply encode the syntactic and semantic relationships within the sentence:
\begin{equation}
    \mathcal{L}_{mlm} = -\sum_{i \in \text{mask}} \log P(w_i | \mathcal{T}_{masked}; \theta_{text})
\end{equation}
By optimizing $\mathcal{L}_{mlm}$, we ensure that the output embeddings used for MaxSim are not merely isolated word representations but are enriched with global context. Note that we apply MLM exclusively to the text encoder rather than to both modalities. This design is motivated by the same asymmetry underlying our unidirectional MaxSim: since it is the text tokens that query the motion patches, the contextual quality of each token embedding is the primary bottleneck for alignment accuracy.

\subsection{Training Strategy and Loss Function}
\label{sec:loss}
For a batch of $B$ text-motion pairs, we compute the pairwise similarity matrix $\mathbf{S}_{\text{batch}} \in \mathbb{R}^{B \times B}$ using the T2M MaxSim Score (Eq.~\ref{eq:maxsim}). The T2M retrieval loss applies in-batch cross-entropy with the diagonal as the target:
\begin{equation}
    \mathcal{L}_{t2m} = -\frac{1}{B}\sum_{i=1}^{B} \log \frac{\exp(\mathbf{S}_{\text{batch}}[i,i] / \tau)}{\sum_{j=1}^{B} \exp(\mathbf{S}_{\text{batch}}[i,j] / \tau)}
\end{equation}
where $\tau$ is a learnable temperature. The M2T loss $\mathcal{L}_{m2t}$ applies the same formulation on the transpose $\mathbf{S}_{\text{batch}}^\top$. The total objective combines the symmetric retrieval loss with the MLM regularization:
\begin{equation}
    \mathcal{L}_{ret} = \tfrac{1}{2}(\mathcal{L}_{t2m} + \mathcal{L}_{m2t}), \qquad
    \mathcal{L}_{total} = \mathcal{L}_{ret} + \lambda_{mlm} \mathcal{L}_{mlm}
\end{equation}
where $\lambda_{mlm}$ balances retrieval and regularization. During training, clean text is used for $\mathcal{L}_{ret}$ and masked text for $\mathcal{L}_{mlm}$; both objectives share the text encoder parameters.

\section{Experiments}
\label{sec:experiments}

\subsection{Datasets and Implementation Details}
We evaluate our proposed method on two datasets. \textbf{HumanML3D}~\cite{guo2022generating} aggregates motion skeleton data from AMASS~\cite{mahmood2019amass} and HumanAct12~\cite{guo2020action2motion}, comprising 14,616 motion clips and 44,970 text descriptions covering diverse daily activities. We follow the standard split~\cite{guo2022generating,petrovich2023tmr}: 23,384/1,460/4,380 for training/validation/testing at 20 FPS~\cite{remogpt2025,petrovich2023tmr,car2024,yu2024exploring}. \textbf{KIT-ML}~\cite{plappert2016kit} contains 3,911 motions with 6,278 descriptions, split into 4,888/300/830 for training/ validation/ testing at 12.5 FPS~\cite{remogpt2025,petrovich2023tmr,car2024,yu2024exploring}.

The system performs T2M by ranking gallery motions given a text query, and M2T by ranking texts given a motion. We report Recall@K (R@K) and Median Rank (MedR) for both T2M and M2T retrieval; higher R@K and lower MedR are better. \new{We also report R@1/R@2, but they} suffer high cross-seed variance due to near-duplicate/similar-motion interference\new{; a 5-seed significance test (Supplementary Material~3) shows that R@5/R@10/MedR separate methods far more reliably, so we treat them as the primary metrics. All results} in Table~\ref{tab:comparison_both_datasets} \new{use} the same seed 42 to ensure fair comparison with previous works. Our base model, which we name \textbf{MotionColBert}, uses ViT-Base and DistilBERT following prior work. To assess scalability, we also report a larger variant, \textbf{MotionColBert-L}, with ViT-Large~\cite{dosovitskiy2020image} and RoBERTa-Large~\cite{liu2019roberta}. For fairness, we compare against MoPatch-L and SGAR-L, which scale MoPatch~\cite{yu2024exploring} and SGAR~\cite{sgar2025} to the same ViT-Large and RoBERTa-Large backbones.

Both encoders project their outputs into a shared latent space with a dimension of $d=256$. We train for 60 epochs on a single NVIDIA H200 GPU with batch size 128, using AdamW~\cite{loshchilov2017decoupled} (weight decay $0.1$) with learning rates of $1{\times}10^{-4}$ and $1{\times}10^{-5}$ for HumanML3D and KIT-ML, respectively. We mask 15\% of tokens for the MLM loss with $\lambda_{\text{mlm}}{=}0.2$.

\subsection{Performance}

\subsubsection{Comparison with State-of-the-Art.}

\begin{table}[t]
\centering
\caption{Comparison with state-of-the-art methods on HumanML3D and KIT-ML datasets. MotionColBert-L uses ViT-Large and RoBERTa-Large backbones. $\dagger$: methods used extra LLM augmentation as input. \new{Best results are in \textbf{bold} and second-best are \underline{underlined}; ``--'': not reported.}}
\label{tab:comparison_both_datasets}
\resizebox{\textwidth}{!}{
\begin{tabular}{l l c cccccc | cccccc}
\toprule
\multirow{2}{*}{\textbf{Dataset}} & \multirow{2}{*}{\textbf{Methods}} & \multirow{2}{*}{\textbf{Venue}} & \multicolumn{6}{c|}{\textbf{T2M Retrieval}} & \multicolumn{6}{c}{\textbf{M2T Retrieval}} \\
& & & \new{R@1$\uparrow$} & \new{R@2$\uparrow$} & R@3$\uparrow$ & R@5$\uparrow$ & R@10$\uparrow$ & MedR$\downarrow$ & \new{R@1$\uparrow$} & \new{R@2$\uparrow$} & R@3$\uparrow$ & R@5$\uparrow$ & R@10$\uparrow$ & MedR$\downarrow$ \\
\midrule
\multirow{12}{*}{\textbf{HML3D}}
& TMR~\cite{petrovich2023tmr} & ICCV'23 & \new{8.92} & \new{12.04} & 16.33 & 22.06 & 33.37 & 25.00 & \new{9.44} & \new{11.84} & 16.90 & 22.92 & 32.21 & 26.00 \\
& CAR~\cite{car2024} & ECCV'24 & \new{10.10} & \new{13.50} & 18.96 & 26.42 & 37.28 & 19.50 & \new{11.18} & \new{13.85} & 20.19 & 26.65 & 36.37 & 20.50 \\
& KinMo$^{\dagger}$~\cite{kinmo2025} & ICCV'25 & \new{9.05} & \new{15.23} & 20.47 & 28.62 & 41.60 & 16.00 & \new{9.01} & \new{15.92} & 21.42 & 29.50 & 41.43 & 16.00 \\
& PL-TMR$^{\dagger}$~\cite{remogpt2025} & AAAI'25 & \new{11.00} & \new{17.02} & 22.18 & 29.48 & 43.43 & \new{\underline{14.00}} & \new{12.25} & \new{14.95} & 21.45 & 28.34 & 39.11 & 19.00 \\
& Lyu et al.~\cite{cmmm2025} & ICLR'25 & \new{11.80} & \new{17.11} & 23.25 & 30.81 & 43.36 & \new{\underline{14.00}} & \new{12.39} & \new{15.55} & 22.17 & 29.25 & 40.34 & 17.00 \\
& MoPatch~\cite{yu2024exploring} & CVPR'24 & \new{11.52} & \new{15.78} & 21.94 & 29.15 & 40.78 & 17.00 & \new{11.70} & \new{14.67} & 21.94 & 29.08 & 38.73 & 18.00 \\
& MoPatch-L & - & \new{12.07} & \new{16.88} & 22.88 & 30.36 & 42.36 & 15.00 & \new{12.29} & \new{15.65} & 22.51 & 29.36 & 40.10 & 17.00 \\
& SGAR$^{\dagger}$~\cite{sgar2025} & NeurIPS'25 & \new{\new{\underline{12.86}}} & \new{--} & 23.52 & 30.75 & 43.00 & 15.00 & \new{\textbf{13.82}} & \new{--} & \new{\underline{23.38}} & 30.09 & \new{\underline{41.83}} & 16.00 \\
& SGAR-L$^{\dagger}$ & - & \new{11.27} & \new{16.43} & 21.59 & 28.50 & 41.37 & 16.00 & \new{11.07} & \new{14.10} & 20.84 & 27.38 & 39.37 & 17.00 \\
& Multi-Modal MR$^{\dagger}$~\cite{multimodalmr2026} & TMM'26 & \new{8.78} & \new{--} & 19.19 & 27.10 & 40.69 & 16.00 & \new{9.32} & \new{--} & 20.78 & 27.64 & 41.61 & \new{\underline{15.00}} \\
\cmidrule{2-15}
& \textbf{MotionColBert} & Proposed & \new{11.87} & \new{\new{\underline{18.13}}} & \new{\underline{23.53}} & \new{\underline{31.71}} & \new{\underline{43.80}} & \new{\underline{14.00}} & \new{13.10} & \new{\new{\underline{16.21}}} & 22.82 & \new{\underline{30.45}} & 41.45 & 16.00 \\
& \textbf{MotionColBert-L} & Proposed & \new{\textbf{13.76}} & \new{\textbf{19.18}} & \textbf{26.22} & \textbf{34.89} & \textbf{48.08} & \textbf{11.00} & \new{\new{\underline{13.76}}} & \new{\textbf{17.42}} & \textbf{25.47} & \textbf{33.22} & \textbf{44.74} & \textbf{13.00} \\
\midrule
\multirow{9}{*}{\textbf{KIT-ML}}
& TMR~\cite{petrovich2023tmr} & ICCV'23 & \new{10.05} & \new{13.87} & 20.74 & 30.03 & 44.66 & 14.00 & \new{11.83} & \new{13.74} & 22.14 & 29.39 & 38.55 & 16.00 \\
& CAR~\cite{car2024} & ECCV'24 & \new{13.87} & \new{19.63} & 25.78 & 37.22 & 53.01 & 11.00 & \new{13.01} & \new{14.69} & 24.93 & 32.53 & 46.74 & 14.00 \\
& Lyu et al.~\cite{cmmm2025} & ICLR'25 & \new{15.13} & \new{\textbf{23.74}} & \new{\underline{31.61}} & 36.81 & 54.12 & \new{\underline{8.00}} & \new{\new{\underline{15.01}}} & \new{\textbf{19.47}} & \textbf{30.06} & 35.63 & 47.53 & \new{\underline{10.50}} \\
& MoPatch~\cite{yu2024exploring} & CVPR'24 & \new{--} & \new{--} & 28.60 & 34.33 & 50.78 & 10.00 & \new{--} & \new{--} & 26.14 & 34.70 & 44.23 & 13.00 \\
& MoPatch-L & - & \new{--} & \new{--} & 28.08 & 33.89 & 50.31 & 10.00 & \new{--} & \new{--} & 25.92 & 34.07 & 44.60 & 13.00 \\
& SGAR$^{\dagger}$~\cite{sgar2025} & NeurIPS'25 & \new{\textbf{16.75}} & \new{--} & 30.48 & 40.00 & 53.61 & 9.00 & \new{14.46} & \new{--} & 25.54 & 35.54 & 48.31 & 11.00 \\
& SGAR-L$^{\dagger}$ & - & \new{\new{\underline{16.39}}} & \new{\new{\underline{23.01}}} & 30.36 & 39.88 & 54.70 & \new{\underline{8.00}} & \new{14.58} & \new{16.87} & 25.90 & \new{\underline{36.02}} & \new{\underline{49.76}} & 11.00 \\
\cmidrule{2-15}
& \textbf{MotionColBert} & Proposed & \new{13.86} & \new{\new{\underline{23.01}}} & \textbf{32.29} & \textbf{43.37} & \textbf{59.28} & \textbf{7.00} & \new{14.46} & \new{16.75} & 25.78 & 35.06 & 49.64 & 11.00 \\
& \textbf{MotionColBert-L} & Proposed & \new{16.27} & \new{22.77} & 30.60 & \new{\underline{42.65}} & \new{\underline{56.39}} & \new{\underline{8.00}} & \new{\textbf{16.02}} & \new{\new{\underline{17.71}}} & \new{\underline{28.80}} & \textbf{38.80} & \textbf{51.33} & \textbf{9.00} \\
\bottomrule
\end{tabular}
}
\end{table}

Table~\ref{tab:comparison_both_datasets} presents a comprehensive comparison with recent methods on HumanML3D and KIT-ML. On both benchmarks, our base model matches or exceeds all methods that do not rely on LLM-based text augmentation, and is competitive with those that do. Notably, on KIT-ML, our base model achieves the best R@10 (59.28\%) and MedR (7.00) in T2M retrieval, outperforming the strongest competing method SGAR-L~\cite{sgar2025} by 4.58 points in R@10, suggesting that fine-grained matching is particularly advantageous when subtle inter-candidate differences must be resolved.

\subsubsection{Scaling to Larger Backbones.}
A key advantage of our framework is its effective scalability. On HumanML3D, MotionColBert-L improves over the base model by +4.28\% in T2M R@10 and reduces MedR from 14 to 11, whereas MoPatch-L and SGAR-L show only marginal or negative gains over their base variants. A similar trend holds on KIT-ML (detailed scaling analysis in Supplementary Material~2.1). These results suggest that global-matching methods suffer from diminishing returns when scaling, as representations are compressed into a single vector. Our structurally decoupled motion representation provides semantically coherent patch features, and the token-level late interaction directly leverages them, allowing additional model capacity to translate into more discriminative matching.

\subsection{Joint-Angle vs.\ Position Score Maps}
\label{sec:scoremap}

\begin{figure}[t]
    \centering
    \includegraphics[width=\linewidth]{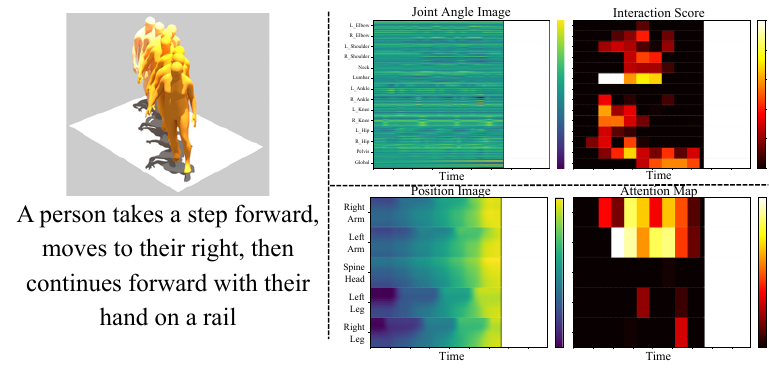}
    \caption{Joint-angle vs.\ position-based representations and their token-to-patch maps. \textbf{Top:} our joint-angle Motion Image (each band is one joint) and its MaxSim interaction score. \textbf{Bottom:} the position-based image (grouped by body region) and its attention map. The joint-angle interaction is sharply localized to the relevant joints and temporal phases, whereas the position-based attention is diffuse.}
    \label{fig:scoremap}
\end{figure}

To make the benefit of our representation concrete, Figure~\ref{fig:scoremap} contrasts the joint-angle Motion Image (ours) with the position-based image (MoPatch~\cite{yu2024exploring}) for the same query, together with the token-to-patch interaction maps. In the joint-angle image every horizontal band encodes a single joint independently, so distinct kinematic events remain separable; in the position image all body-region bands drift together with the global trajectory (the left-to-right colour shift), blurring per-joint detail. By aggregating the per-token, per-patch MaxSim scores into an interaction score map, our score map grounds each phase of motion on the correct joints: it attends to the lumbar and leg bands (L/R knee and hip) during the stepping phase, then shifts to the right shoulder and the global-position row when the hand reaches the rail (reflecting the translational displacement). The position-based attention map is diffuse and concentrated on the arm region, with no clear temporal or part-level correspondence. The joint-angle representation paired with MaxSim thus achieving fine-grained alignment and provides an interpretability that position-based methods lack. Additional score-map and retrieval comparisons are given in Supplementary Material~5.

\subsection{Ablation Study}
Table~\ref{tab:ablation_study} reports a systematic ablation starting from joint positions (Pos) with global matching (Global), incrementally adding: (1) joint angle representation (Angle), (2) MaxSim late interaction, and (3) MLM regularization. We also provide (4) an efficiency analysis. All experiments use the ViT-Large and RoBERTa-Large backbones (our MotionColBert-L configuration). Additional ablations regarding encoder scaling and hyperparameter sensitivity are in Supplementary Material~2.

\begin{table}[t]
\centering
\caption{Ablation study of feature representations, retrieval strategies, and Masked Language Modeling (MLM) on HumanML3D.}
\label{tab:ablation_study}
\resizebox{\textwidth}{!}{
\begin{tabular}{l ccc cccc | cccc}
\toprule
\multirow{2}{*}{\textbf{Dataset}} & \multirow{2}{*}{\textbf{Repr.}} & \multirow{2}{*}{\textbf{Strategy}} & \multirow{2}{*}{\textbf{MLM}} & \multicolumn{4}{c|}{\textbf{T2M Retrieval}} & \multicolumn{4}{c}{\textbf{M2T Retrieval}} \\
& & & & R@3$\uparrow$ & R@5$\uparrow$ & R@10$\uparrow$ & MedR$\downarrow$ & R@3$\uparrow$ & R@5$\uparrow$ & R@10$\uparrow$ & MedR$\downarrow$ \\
\midrule
\multirow{8}{*}{HML3D}
 & Pos   & Global & \ding{55} & 22.88 & 30.36 & 42.36 & 15.00 & 22.51 & 29.36 & 40.10 & 17.00 \\
 & Pos   & Global & \ding{51} & 20.72 & 28.89 & 40.02 & 17.00 & 20.45 & 28.20 & 36.29 & 19.50 \\
 & Pos   & MaxSim & \ding{55} & 23.51 & 30.54 & 43.17 & 15.00 & 22.19 & 29.49 & 40.64 & 17.00 \\
 & Pos   & MaxSim & \ding{51} & 23.88 & 31.00 & 44.18 & 14.00 & 23.20 & 30.13 & 41.76 & 16.00 \\
 & Angle & Global & \ding{55} & 23.63 & 31.11 & 44.72 & 14.00 & 23.67 & 30.21 & 40.92 & 17.00 \\
 & Angle & Global & \ding{51} & 22.57 & 29.60 & 41.08 & 16.00 & 20.90 & 28.08 & 37.39 & 19.00 \\
 & Angle & MaxSim & \ding{55} & 25.03 & 32.39 & 45.64 & 12.00 & 23.64 & 31.67 & 42.30 & 15.00 \\
 & Angle & MaxSim & \ding{51} & \textbf{26.22} & \textbf{34.89} & \textbf{48.08} & \textbf{11.00} & \textbf{25.47} & \textbf{33.22} & \textbf{44.74} & \textbf{13.00} \\
\bottomrule
\end{tabular}
}
\end{table}

\paragraph{Effect of joint angle representation.}
Switching from joint positions to joint angles consistently improves higher-recall metrics under both matching strategies. The improvement is more pronounced under MaxSim (\eg, on HumanML3D T2M R@5 improves by +1.85\% under MaxSim vs.\ +0.75\% under global matching), confirming that decoupling local joint movement from global trajectory produces a cleaner signal for retrieval. This benefit is amplified when combined with fine-grained matching, since MaxSim aligns individual tokens to motion patches, the structured, per-joint nature of the angle representation provides more semantically coherent patch features.

\paragraph{Effect of MaxSim late interaction.}
Replacing global matching with token-level MaxSim improves retrieval under both representations (\eg, T2M R@10 rises from 42.36 to 43.17 for joint positions and from 44.72 to 45.64 for joint angles). This synergy is strongest once MaxSim is further paired with MLM (Angle+MaxSim+MLM reaches 48.08 T2M R@10), validating our design of co-locating fine-grained matching with a structurally decoupled representation.

\begin{table}[t]
\centering
\caption{Efficiency and compression trade-off on HumanML3D. PQ: Product Quantization; Binary: asymmetric binary hashing. Query latency measured on a single NVIDIA H200 GPU.}
\label{tab:efficiency_merged}
\resizebox{\columnwidth}{!}{%
\begin{tabular}{l c c c c c}
\toprule
Method & Storage & Comp. & T2M R@10$\uparrow$ & M2T R@10$\uparrow$ & Latency (ms) \\
\midrule
MoPatch (global)              & 4.28\,MB   & -- & 40.78 & 38.73 & 3.14 \\
\midrule
MotionColBert (Float32)                & 837.21\,MB & 1$\times$  & 43.80 & 41.45 & 4.10 \\
MotionColBert + PQ ($m{=}64$, 8bit)   & 52.32\,MB  & 16$\times$ & 43.39 {\small(\textminus 0.41)} & 41.06 {\small(\textminus 0.39)} & -- \\
MotionColBert + Binary                 & 26.14\,MB  & 32$\times$ & 43.10 {\small(\textminus 0.70)} & 39.87 {\small(\textminus 1.58)} & -- \\
\bottomrule
\end{tabular}%
}
\end{table}

\paragraph{Effect of MLM regularization.}
MLM helps only when combined with token-level MaxSim, and is detrimental under global pooling. With MaxSim, adding MLM improves every metric: on HumanML3D it raises Angle+MaxSim T2M R@10 from 45.64 to 48.08 (+2.44) and M2T R@10 from 42.30 to 44.74 (+2.44), with similar gains for joint positions. In contrast, under global matching the same objective hurts performance (Angle+Global T2M R@10 drops from 44.72 to 41.08, and Pos+Global from 42.36 to 40.02). This asymmetry reinforces the coupled design of our framework: MLM enriches each token embedding with sentence-level context, but this context can only be exploited by a token-level matching operator. A single pooled embedding discards the per-token structure that MLM regularizes, so the auxiliary objective merely competes with the contrastive signal rather than aiding it. Further sensitivity analysis on the MLM loss weight ($\lambda_{mlm}$) is provided in Supplementary Material~2.2.

\paragraph{Efficiency analysis.}
Table~\ref{tab:efficiency_merged} reports the efficiency and compression trade-offs on HumanML3D. Following standard retrieval practice~\cite{khattab2020colbert,faysse2024colpali}, all motion embeddings are pre-computed offline. Since MaxSim retains $N{=}196$ patch embeddings per motion, gallery storage increases to ${\sim}837$\,MB vs.\ ${\sim}4$\,MB for MoPatch's global vectors. However, query latency increases only marginally (4.10\,ms vs.\ 3.14\,ms; on KIT-ML: 3.19\,ms vs.\ 2.79\,ms) due to GPU-parallel similarity computation. Post-hoc compression effectively mitigates storage: Product Quantization (PQ)~\cite{jegou2010product} with $m{=}64$ achieves 16$\times$ compression with ${\leq}0.41\%$ R@10 loss, and binary hashing~\cite{gong2012iterative} at 32$\times$ retains 43.10\% T2M R@10.


\section{Conclusion}
\label{sec:conclusion}
We presented a framework for fine-grained text-motion retrieval that overcomes the information bottleneck of global-embedding methods by combining an anatomically defined joint-angle representation with a MaxSim late-interaction mechanism and MLM regularization. Extensive experiments demonstrate state-of-the-art retrieval performance and interpretable token-to-patch correspondence maps. In practice, this transparency empowers animators to search vast motion databases while verifying specific part-level matches, and provides a robust foundation for downstream tasks like language-driven motion generation and localized editing. While our approach unlocks superior accuracy and interpretability, retaining dense patch embeddings introduces a trade-off in increased storage and computational overhead for offline galleries. Future work will explore efficient indexing strategies, such as approximate nearest-neighbor search and vector quantization, to scale this interpretable retrieval framework to massive, industry-level motion repositories.

{
\medskip\noindent\textbf{Acknowledgements.} This work was a part of Finland's Ministry of Education and Culture's Doctoral Education Pilot under Decision No.~VN/\allowbreak 3137/\allowbreak \mbox{2024-OKM-6} (The Finnish Doctoral Program Network in Artificial Intelligence, \mbox{AI-DOC}).\par
}

\bibliographystyle{splncs04}
\bibliography{main}

\end{document}